\documentclass{article}

\usepackage{PRIMEarxiv}

\usepackage[utf8]{inputenc} 
\usepackage[T1]{fontenc}    
\usepackage{hyperref}       
\usepackage{url}            
\usepackage{booktabs}       
\usepackage{amsfonts}       
\usepackage{nicefrac}       
\usepackage{microtype}      
\usepackage{lipsum}
\usepackage{fancyhdr}       
\usepackage{graphicx}       
\graphicspath{{media/}}     

\title{Can AI Read Between the Lines? Benchmarking LLMs on Financial Nuance
}

\author{
  Dominick Kubica\textsuperscript{a+}, 
  Dylan T. Gordon\textsuperscript{a+}, 
  Nanami Emura\textsuperscript{a+}, 
  Derleen Saini\textsuperscript{a+}, 
  and Charlie Goldenberg\textsuperscript{a+} \\
  \textsuperscript{a}Department of Business Analytics, Santa Clara University - Leavey School of Business,\\
  Santa Clara, California 95053, United States \\
  \textsuperscript{+}These authors contributed equally.\\
  \texttt{\{dkubica, dtgordon, nemura, dsaini, cgoldenberg\}scu@edu}
}

\begin{document}
\maketitle

\vspace{-2em} 

\begin{center}
\small This research was conducted as part of a Microsoft-sponsored Capstone Project at Santa Clara University, led by Juhi Singh and Bonnie Ao from the Microsoft MCAPS AI Transformation Office.
\end{center}

\vspace{2em} 

\begin{abstract}
As of 2025, Generative Artificial Intelligence (GenAI) has become a central tool for productivity across industries. Beyond text generation, GenAI now plays a critical role in coding, data analysis, and research workflows. As large language models (LLMs) continue to evolve, it is essential to assess the reliability and accuracy of their outputs, especially in specialized, high-stakes domains like finance. Most modern LLMs transform text into numerical vectors, which are used in operations such as cosine similarity searches to generate responses. However, this abstraction process can lead to misinterpretation of emotional tone, particularly in nuanced financial contexts. While LLMs generally excel at identifying sentiment in everyday language, these models often struggle with the nuanced, strategically ambiguous language found in earnings call transcripts. Financial disclosures frequently embed sentiment in hedged statements, forward-looking language, and industry-specific jargon, making it difficult even for human analysts to interpret consistently, let alone AI models. This paper presents findings from the Santa Clara Microsoft Practicum Project, led by Professor Charlie Goldenberg, which benchmarks the performance of Microsoft’s Copilot, OpenAI’s ChatGPT, Google’s Gemini, and traditional machine learning models for sentiment analysis of financial text. Using Microsoft earnings call transcripts, the analysis assesses how well LLM-derived sentiment correlates with market sentiment and stock movements and evaluates the accuracy of model outputs. Prompt engineering techniques are also examined to improve sentiment analysis results. Visualizations of sentiment consistency are developed to evaluate alignment between tone and stock performance, with sentiment trends analyzed across Microsoft’s lines of business to determine which segments exert the greatest influence.
\end{abstract}

\section{Introduction}
Generative AI’s role in high-stakes domains like finance will become more prevalent as AI becomes increasingly embedded in professional workflows. Financial language is uniquely complex because it is charged with forward-looking statements, hedged language, and subtle cues that challenge current models. Can today’s Large Language Models (LLMs) understand this kind of nuance? \\
This question motivated a collaborative research project between Santa Clara University and the Microsoft data science team. The evaluation centers on whether LLMs can outperform traditional natural language processing (NLP) tools in financial sentiment analysis and whether they can generate useful insights when applied to real-world financial reporting such as quarterly earnings calls.\\
The approach had three parts:
\begin{enumerate}
\item Benchmarking LLMs and traditional NLP tools on a standardized financial dataset.
\item Applying these models to Microsoft’s quarterly earnings transcripts \cite{microsoftInvestor2025}, breaking down sentiment by business line, and better understanding insights that can be extracted from earnings call transcripts. 
\item Analyzing results to identify optimization opportunities and assess how sentiment correlates with actual stock performance.
\end{enumerate}

The results were both encouraging and eye-opening: while LLMs significantly outperformed traditional tools in grasping nuanced sentiment, they still face performance challenges.  This paper outlines the benchmarking process, real-world findings, and recommendations to enhance tools like Microsoft Copilot.

\section{Evaluating the Accuracy of Models Through Benchmarking}
\label{sec:headings}

An objective benchmarking process is essential to evaluate performance differences between LLMs and traditional NLP tools.  A standardized evaluation was conducted to measure how accurately various models interpret sentiment in financial texts. Given the complexities of financial language, this comparison highlights how effectively each model captures tone and nuance, offering insights for both tool selection and future model development. 
Accuracy testing was conducted using the Financial Phrase Bank dataset, developed by researchers at Aalto University. \cite{sbhatti2021financial} It consists of financial and earnings-related news headlines labeled as positive, neutral, or negative based on market sentiment. Nine models were compared: 

\begin{itemize}
\item LLM-based/cloud platforms: Microsoft Copilot Desktop App, Copilot via Microsoft 365, Copilot App Online\footnote{ChatGPT is developed by OpenAI and operates on Microsoft’s Azure supercomputing infrastructure. While Microsoft and OpenAI collaborate in the development and delivery of AI services, OpenAI remains an independent entity. Azure OpenAI Service provides enterprise-grade access to OpenAI models.}, ChatGPT - 4o, and Google Gemini 2.0 Flash 
\item Cloud-based NLP service: Azure Language AI  
\item Python libraries: FinBERT (Transformer model via Python library), NLTK, and TextBlob (Microsoft Copilot 365)
\end{itemize}

Each model classified the same sentences from the dataset. Financial sentences were preprocessed for the traditional NLP libraries to ensure formatting consistency. For LLM-based tools, identical prompts were used to reflect a real-world application. After each model returned the sentiment of each sentence, accuracy was measured as the percentage of correct classifications against the pre-labeled dataset. Both the Copilot desktop app and the Chat interface, whether run locally or accessed via the web, were used with the ''Think Deeper'' capability. Microsoft 365 does not have a ''Think Deeper'' ability available.

\begin{figure}
    \centering
    \includegraphics[width=0.75\linewidth]{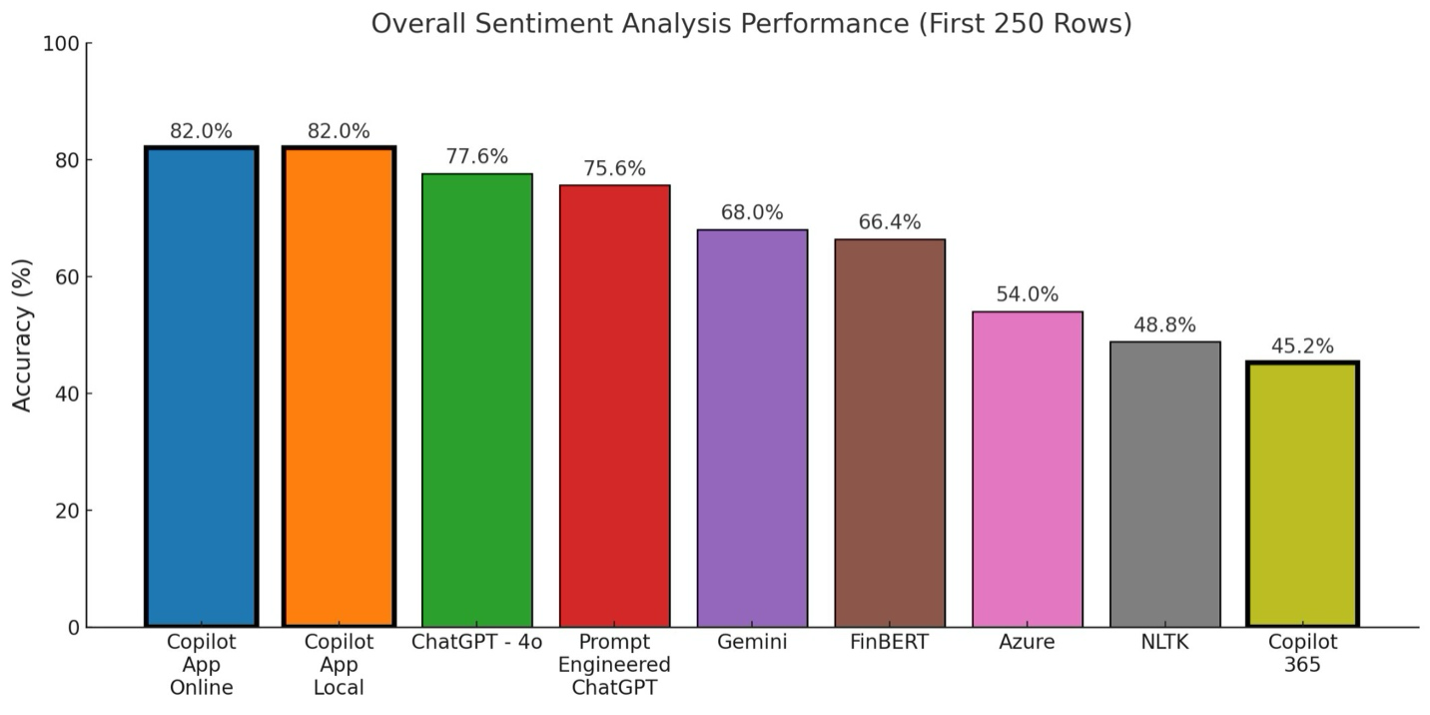}
    \caption{Overall Sentiment Analysis Performance (First 250 Rows)}
    \label{fig:fig1}
\end{figure}

Benchmarking revealed significant differences in sentiment analysis accuracy across models (Figure \ref{fig:fig1}). The Copilot App (both Online and Local) led with an accuracy of 82.0\%, followed by ChatGPT 4o (77.6\%), Prompt-Engineered ChatGPT (75.6\%), and Gemini (68.0\%). Notably, LLMs using uncleaned sentences demonstrated stronger performance, particularly in detecting nuance and hedged expressions.  

\begin{figure}
    \centering
    \includegraphics[width=0.75\linewidth]{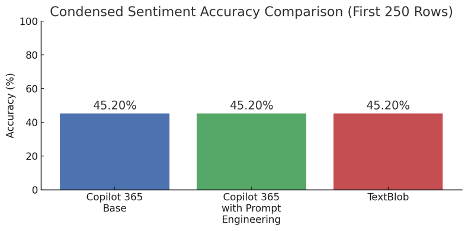}
    \caption{Condensed Sentiment Accuracy Comparison (First 250 Rows)}
    \label{fig:fig2}
\end{figure}

Copilot through Microsoft 365 exhibited lower accuracy compared to other LLM-based sentiment models in the benchmarking. Analysis indicated that Copilot 365 defaults to the TextBlob Python library as its primary sentiment analysis engine, while both the desktop App and web-hosted Chat versions utilize their full LLM-based capabilities. Both Prompt Engineered and Unguided Copilot 365 returned outputs consistent with standalone TextBlob, often defaulting to neutral sentiment and missing implied or domain-specific cues (Figure \ref{fig:fig2}). This outcome is reflective of the design intent of Copilot 365, which is primarily focused on enhancing the productivity features of the Microsoft 365 suite rather than advanced NLP tasks. \\
Handling CSV files presented challenges for Copilot across all tested versions. The models occasionally struggled to interpret structured data accurately, and converting CSVs to plain text was often necessary to improve reliability. Instances of hallucination were observed during formatting and post-processing, resulting in inconsistent outputs. In cases where Copilot defaulted to simpler tools or miscommunicated its capabilities, the lack of transparency risked eroding user trust. This is especially sensitive domains like financial analysis, where clarity and reliability are critical.\\
These findings highlight that deployment choices, including model selection, tool integration, and input formatting, have a significant impact on LLM performance in specialized tasks. In contexts like financial sentiment analysis, these factors can determine whether outputs are accurate, reliable, and useful. Across the board, LLMs outperformed traditional sentiment engines in identifying implied or nuanced sentiment. ChatGPT and Gemini delivered strong results, while FinBERT was particularly effective for finance-specific cases. The performance of the Copilot App illustrates the potential of integrated LLM tools for financial analysis, contingent upon appropriate model selection and deployment.

\section{Real-World Application: Business Line Sentiment vs Stock Prediction}

While benchmarking LLMs on a standardized dataset offered valuable insight into model accuracy, it was also important to assess whether these tools could convert benchmarking results into actionable insights in a real-world financial context. Specifically, can business line sentiment derived from earnings call transcripts provide additional business insights or competitive data, and correlate with stock price movements?\\
Microsoft Copilot was used to segment quarterly earnings call transcripts by business line: Devices, Dynamics, Gaming, Office Commercial, Search and News Advertising, Server Products and Cloud Services, Q and A, and Overall Sentiment. Each segment was then processed using ChatGPT to evaluate sentiment at the business-line level. ChatGPT-4o was selected as the sentiment analysis tool due to its consistent output and its ability to process multiple quarters of transcripts efficiently through a single standardized Python workflow. While Copilot demonstrated the highest sentiment accuracy in our evaluation, its lack of API access and reliance on manual input made it impractical for large-scale analysis. Given ChatGPT-4o’s close performance and superior accessibility, it was the most viable choice for the analysis.

\begin{figure}
    \centering
    \includegraphics[width=1\linewidth]{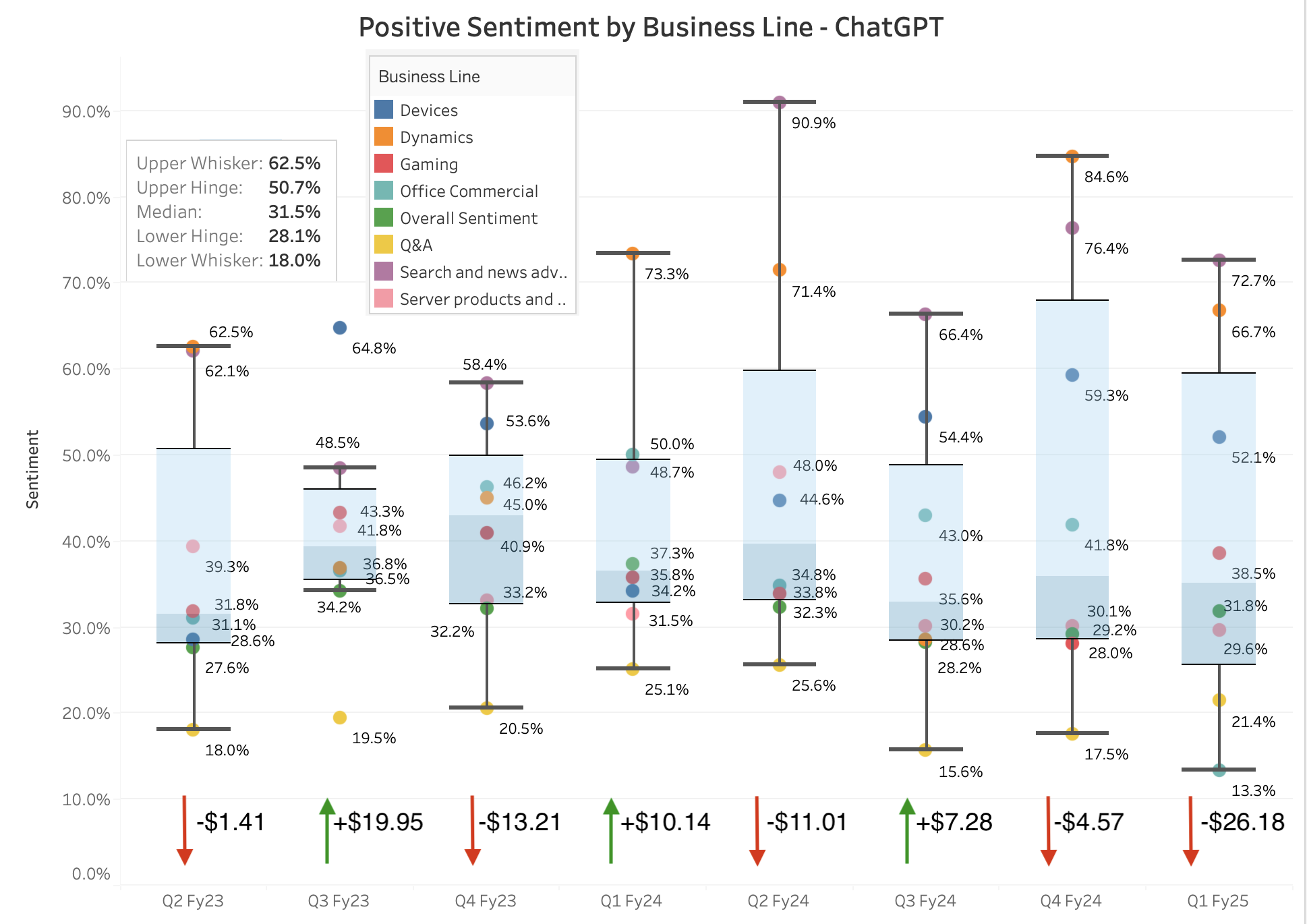}
    \caption{Positive Sentiment by Business Line - ChatGPT}
    \label{fig:fig3}
\end{figure}

The blue shaded box for each quarter shows the interquartile range of the data and the black horizontal lines represent the minimum and maximum sentiment values without outliers (Figure \ref{fig:fig3}). The arrows below represent the stock price increase or decrease the day after the earnings call for each quarter.\\
Analyzing overall transcript sentiment provided limited insight into stock movement, but breaking down sentiment by business segment revealed meaningful patterns that would otherwise be overlooked. For instance, high positive sentiment in the ''Search and News Advertising'' segment during Q1 2025 was associated with a notable drop in stock price following the call. Conversely, in Q3 2023, Devices sentiment spiked positively and was followed by a significant rise in Microsoft’s share price. These findings suggest that sentiment within specific business segments may have a greater impact on market response than the overall tone of each earnings call.

\begin{figure}
    \centering
    \includegraphics[width=0.75\linewidth]{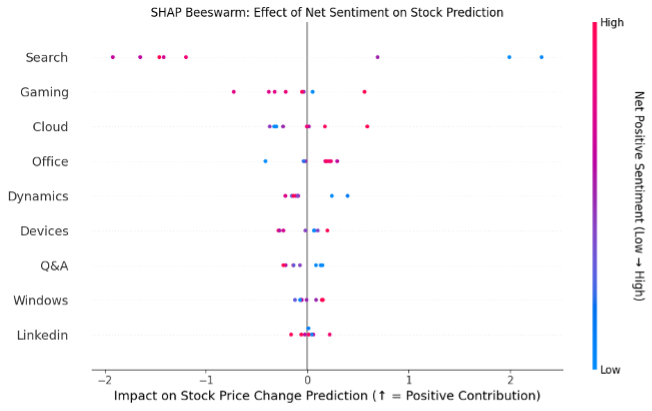}
    \caption{SHAP Beeswarm: Effect of Net Sentiment on Stock Prediction}
    \label{fig:fig4}
\end{figure}

To visualize this relationship, the above SHAP (SHapley Additive exPlanations) beeswarm plot was created to map sentiment direction against stock movement (Figure \ref{fig:fig4}). Each dot represents a business segment within a specific quarter. Red indicates high positive sentiment, while blue indicates low sentiment. The horizontal axis represents the SHAP value, indicating the contribution of each business line’s sentiment to the model’s stock price change prediction, with values further from zero signifying greater influence. \\
Importantly, dots for Search and News Advertising cluster toward the left despite the red color. This reinforces the hypothesis that a positive tone in this segment may raise investor skepticism or signal over-optimism, resulting in stock price decline.\\
This inverse correlation raises a critical insight: not all positive sentiment within transcripts translates to positive external or investor sentiment.  Similar sentiment inversions were observed in segments like gaming and Q\&A as well, where optimistic statements were followed by negative stock movement. These cases illustrate a core challenge in sentiment analysis: tone alone is not a reliable predictor of investor sentiment. Investor response likely depends on additional context, expectations, and broader market narratives.\\
Ultimately, breaking down transcripts by business line was key to revealing meaningful insights and it transformed high-level analysis into more granular interpretation. Integrating LLMs into workflows offers a valuable opportunity to improve forecasting accuracy and contextual understanding. When combined with human expertise and domain knowledge, LLMs can uncover nuanced sentiment patterns that open avenues for further exploratory analysis and research. The findings indicate that LLMs, when combined with domain expertise, can support the identification of sentiment patterns that may inform further analysis, while acknowledging that conclusive stock movement predictions remain outside the current capabilities of these approaches.

\section{Findings and Optimization Recommendations}
Reviewing our benchmark results alongside the Microsoft earnings case study, several key insights emerged:

\begin{itemize}
\item LLMs clearly outperformed traditional tools in financial sentiment analysis, especially in detecting filler words and subtle cues.
\item Traditional models require aggressive text cleaning, which often strips away nuance. LLMs were able to use more of the filler context wording.
\item Despite their edge, LLMs still failed to exceed 85\% accuracy. Financial experts should be able to surpass this with their tailored domain knowledge.
\item LLMs remain expensive and computationally intensive, limiting scalability for smaller teams.
\item Human creativity is still essential. While LLMs supported tasks like code generation and visualizations, they lacked the intuition to guide the project’s direction. The key questions, analysis choices, and meaningful visualizations came from the data scientists. AI can assist, but it doesn’t have the perspective or creative insight to see a project through from start to finish.
\end{itemize}

These findings reinforce that LLMs are powerful tools, but not replacements for domain experts.

\subsection{Observed Optimization Areas in Copilot and LLM Implementations for Financial Applications}
Evaluation of multiple LLM platforms, including different Copilot implementations, identified several areas for potential optimization to enhance performance and usability in financial sentiment analysis tasks.

\begin{itemize}
\item Performance Transparency:\\
In some configurations, tasks were routed to traditional tools like TextBlob rather than handled directly by the LLM, but users were not clearly informed when this occurred.  Clear indicators of fallback behavior can help users understand what the system is doing. In addition, documentation that explains capability differences between configurations supports more informed and confident decision-making. Without this transparency, users may misinterpret the system’s capabilities, leading to confusion and reduced confidence in the tool. Over time, this lack of clarity can erode user trust, especially in professional or high-stakes applications where reliability is critical.
\item Structured Data Handling:\\
All LLM systems should prioritize reliable handling of structured data formats like CSVs. In Copilot, accurate analysis often required converting CSVs to plain text, adding extra steps for the user. While some internal processing may be needed, this complexity should be managed by the system, not the user. Streamlining CSV support can improve usability in data-heavy workflows and encourage adoption in professional settings.
\item Reliability in Basic NLP Functions:\\
Hallucinations and inconsistencies in core tasks, such as sentiment counting and text cleaning, were frequent and deeply concerning. In some cases, LLMs fabricated capabilities or produced results that were clearly inaccurate, even when the task was straightforward. These failures are concerning in specialized domains where accuracy is non-negotiable. If left unaddressed, they will continue to undermine user confidence and limit the practical adoption of LLMs in high-stakes settings.
\end{itemize}

\section{Conclusion}
Financial language is layered with strategy, hedging, and nuance, making it a stress test for any sentiment analysis tool. This study found that while LLMs outperform traditional NLP libraries in detecting financial sentiment, they still face architectural, economic, and reliability barriers to adoption at scale.\\
When used carefully, however, LLMs can help highlight patterns or shifts that may complement traditional analysis, especially when applied at the business-line level.  The Microsoft earnings call case study demonstrates that a segmented, model-enhanced approach can connect executive tone with investor behavior in powerful ways.\\
Looking forward, tools like Microsoft Copilot have immense potential, but unlocking that potential will require deeper integration with LLM capabilities, better transparency, and a stronger focus on enterprise needs.\\
LLMs, paired with human intelligence, will reshape the future of financial analysis not by replacing experts, but by providing them with a powerful, new toolset.

\begingroup
\raggedright
\bibliographystyle{unsrt}  
\bibliography{references}
\endgroup

\end{document}